\documentclass[letterpaper, 10 pt, conference]{ieeeconf}  

\IEEEoverridecommandlockouts                              

\title{\LARGE \bf
A Framework for Realistic Simulation of Daily Human Activity
}

\usepackage[numbers,sectionbib]{natbib}
\usepackage{color, soul}
\usepackage{graphicx}
\usepackage{amsmath}
\usepackage{mathtools}
\usepackage{algorithm}
\usepackage{algpseudocode}
\usepackage{algorithmicx}
\usepackage{amssymb}
\usepackage[table]{xcolor}

\newif\ifpaperfinal
\paperfinalfalse

\ifpaperfinal
\newcommand{\iinote}[1]{}
\newcommand{\dgnote}[1]{}
\else
\newcommand{\iinote}[1]{\textcolor{orange}{\textbf{Ifrah: #1}}}
\newcommand{\dgnote}[1]{\textcolor{red}{\textbf{Dylan: #1}}}
\fi
\usepackage{url}

\author{Ifrah Idrees$^{1,*}$, Siddharth Singh$^{2}$, Kerui Xu$^{2}$, Dylan F. Glas$^{2}$
\thanks{$^{1}$Ifrah Idrees is affiliated with Dept. of Computer Science, Brown University, Providence, RI, USA {\tt\small ifrah\_idrees@brown.edu}}%
\thanks{$^{2}$Siddharth Singh, Kerui Xu, and Dylan F. Glas are affiliated with Amazon Lab126, 1100 Enterprise Way, Sunnyvale, CA, USA {\tt\small \{hartsid, xkerui, dgglas\}@amazon.com}}%
~\thanks{$^{*}$Work done during an Amazon Lab126 internship.}%
}

\begin{document}

\maketitle
\thispagestyle{empty}
\pagestyle{empty}



\begin{abstract}
For social robots like Astro which interact with and adapt to the daily movements of users within the home, realistic simulation of human activity is needed for feature development and testing. This paper presents a framework for simulating daily human activity patterns in home environments at scale, supporting manual configurability of different personas or activity patterns, variation of activity timings, and testing on multiple home layouts. We introduce a method for specifying day-to-day variation in schedules and present a bidirectional constraint propagation algorithm for generating schedules from templates. We validate the expressive power of our framework through a use case scenario analysis and demonstrate that our method can be used to generate data closely resembling human behavior from three public datasets and a self-collected dataset. Our contribution supports systematic testing of social robot behaviors at scale, enables procedural generation of synthetic datasets of human movement in different households, and can help minimize bias in training data, leading to more robust and effective robots for home environments.

\end{abstract}

\section{Introduction}

Development of a commercial robot to coexist with people over the long term in a home environment is a challenging task. For robot behaviors which depend on spatial interaction with users, prediction of their locations, or long-term adaptation to user behavior, it can be difficult to evaluate performance effectively through on-device testing alone.

\begin{figure}
    \centering
    \includegraphics[width=\linewidth]{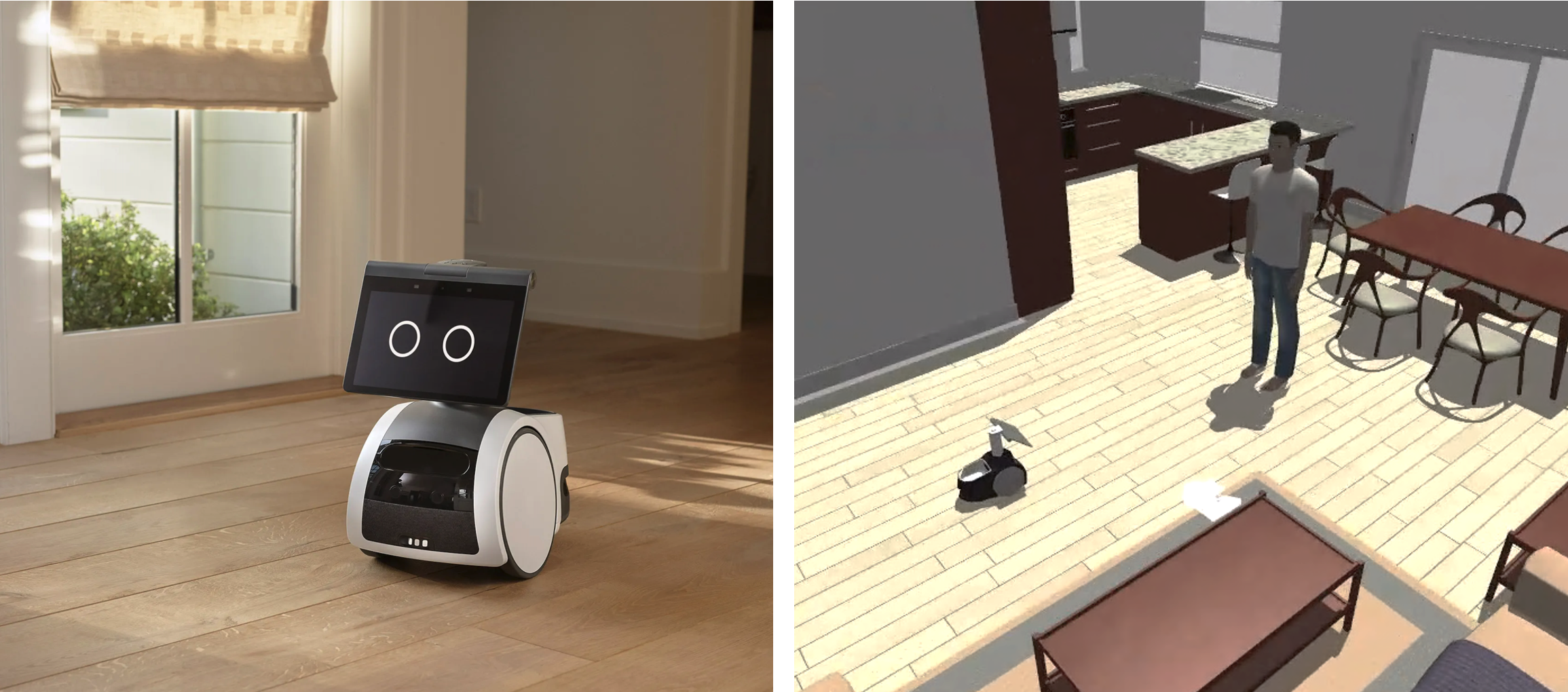}
    \caption{Left: Amazon's Astro robot.  Right: Gazebo-based simulation environment used in feature development and testing for Astro.}
    \label{fig:astro}
\end{figure}

To support the development of behaviors like these for Amazon's Astro robot, we developed a simulation framework for scalable generation of daily human activity, enabling robot testing in a simulation environment with humans moving realistically throughout the home (Fig. \ref{fig:astro}). This approach is not specific to Astro, and could be used for any social robot or smart-home system designed for long-term deployment.

\textbf{User simulation}
Simulation testing is needed when robot capabilities depend on user behavior patterns. For instance, if the robot is asked to deliver a message to a user and needs to search the home for them, simulation testing is needed to quantify the efficiency of the search and its success rate.  

To be effective, these tests should be performed in a variety of simulated homes, at various times of day, and with realistic user behavior patterns. Regardless of whether the robot is meant to learn and adapt to user behaviors or follow fixed rules, realistic testing is necessary to robustly evaluate its effectiveness before it can be released to real users.

\textbf{Avoiding bias in training data}
There has recently been a great deal of social awareness around the topic of bias in AI, both in popular media and in social robotics research \cite{hitron2023implications, parreira2023did, foster2023social}. While there are complex and subtle aspects to this topic, one common type of AI bias occurs when the training data are not representative of the eventual users of a system.

In the context of the current work, if the testing scenarios were all based on large homes with spacious floorplans, the robot might not perform well in smaller homes. Similarly, if the user behavior patterns in testing scenarios only represented working couples with children, the robot might not perform well in the home of a single retiree living alone.

To avoid this kind of bias, it is important to ensure that testing incorporates a diversity of floorplans and simulated user behaviors representing a broad range of possible users. Note that this work focuses on spatial and timing aspects of human activity but not computer vision, so we do not directly address common bias-related topics related to race, gender or skin tone.

\textbf {Manual tuning}
For a commercial product, systematic testing is essential for prevention of regressions as features are developed over time. Quality Assurance (QA) teams develop thorough testing plans to cover both typical usage patterns and any anticipated edge case conditions which might cause failures. For this reason, the ability to hand-craft and tune test cases is important, providing QA with the ability to precisely specify user behavior for each test case.

\textbf{Requirements}
The objective of this simulation framework is thus to satisfy the following requirements:
\begin{enumerate}
    \item Manual control over simulated behaviors
    \item Configurability for different personas or lifestyles
    \item Day-to-day variation of activity timings
    \item Execution on a variety of different floorplans
\end{enumerate}

This paper presents a novel framework for generating configurable and variable schedules for daily human activity satisfying the above requirements, enabling testing of robot behaviors at scale in a simulation environment.

\section{Related Work}
\subsection{Simulating human behavior}
Simulation of human behavior is not new in HRI, and it is often an important element of social navigation research. Variants of the social force model \cite{helbing1995social} are often used for modeling crowds and multi-person scenarios. Systems such as SocNavBench \cite{Biswas2022SocNavBench} and SEAN \cite{Tsoi_2020_HAI} use playback of prerecorded trajectory patterns based on real data as a basis for evaluating robot navigational behavior. Kidokoro et al. used models of observed pedestrian subgoals and stochastically generated pedestrians to follow those routes \cite{kidokoro2015simulation}, and Kaneshige et al. found that using simulation of pedestrian behavior in testing greatly improved the efficiency of robot behavior development in a shopping mall \cite{kaneshige2021overcome}.

These approaches all focus on short-term or local behavior, often with anonymous pedestrians who appear and disappear. Our focus in this work is to model longer-term behavior of a persistent user (or users) in the home throughout the day.  

\subsection{Human schedule generation from data}
Some work has focused on building generative models based on captured human behavior data.
Francillette et al. developed a method for generating behavior trees from historical activity data \cite{francillette2020modeling}. However, this work focused on granular actions such as manipulation of cups and dishes, whereas our interest is in longer-term movement around the home. Elbayoudi et al. \cite{elbayoudi2015modelling} proposed a system for simulating activities of daily living for a targeted population of older adults based on captured data using Hidden Markov Model and Direct Simulation Monte Carlo methods, which, like our work, was applied at the level of room-to-room transitions.

Patel et al. built generative models of daily life behavior from crowdsourced activity schedules, using Gaussian mixture models to capture average times and variability of daily activities \cite{patel2022proactive} but did not model sequence dependencies or hard time constraints. Garcia-Ceja et al. \cite{garcia2014long} modeled daily life activities using conditional random fields. However, their focus was on estimation of human activity given accelerometer data, rather than generation of synthetic data.

None of the models described above were intended to be manually adjusted, whereas a primary aim of our work is to provide designers and testers with manual control over behaviors, although the ability to simulate behavior patterns which imitate observed data is also of interest to us.

\subsection{Existing Human Simulators}

Virtual Home \cite{https://doi.org/10.48550/arxiv.1806.07011} and Alfred \cite{ALFRED20} work well in generating probabilistic human motion patterns referred to as activity programs over multi-step, short-term  tasks, but do not focus on modeling activity variation from day to day.  To our knowledge, there does not exist work that focuses on generating varying daily activity schedules for different user profiles.

\section{Our Approach}

\begin{figure}[thb]
    \centering
    \includegraphics[width=\linewidth]{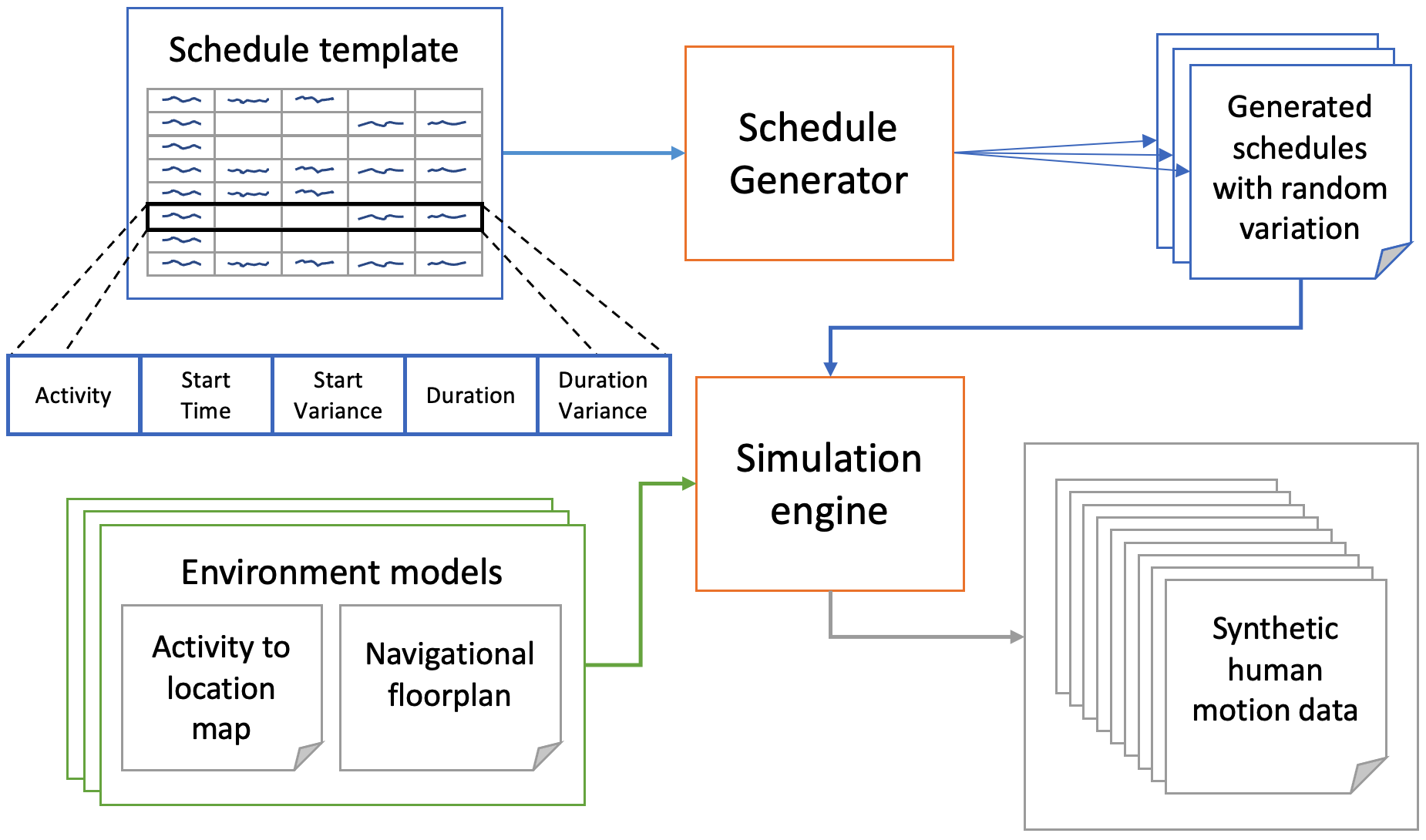}
    \caption{Elements of the proposed simulation framework}
    \label{fig:system-diagram-1}
\end{figure}

The elements of our proposed framework are shown in Fig. \ref{fig:system-diagram-1}, A designer first creates a \textbf{schedule template} specifying a time sequence of abstracted activities for a simulated user's day, including optional time constraints and ranges of variability for start times and activity durations.  The \textbf{schedule generator} processes this template to generate a set of varying daily activity schedules.
To support simulation in different homes, several \textbf{environment models} are defined, each containing a  \textbf{navigational floorplan}, used for path planning in the simulator, and a corresponding \textbf{activity-to-location map} defining specific ($x,y$) coordinates for each activity. 
The generated schedules are executed by the \textbf{simulation engine} to move simulated users around the home according to their scheduled activities throughout the day, generating a diverse set of synthetic data scenarios containing day-to-day schedule variations and a variety of home environments.

Activities are mapped to specific locations in the home.  For example, $cooking \rightarrow kitchen$. For each floorplan, we require an ($x,y$) location to be defined for each activity, allowing the floorplans to be interchanged, so the same schedule template can be used in different home layouts.

As the path-planning and obstacle-avoidance aspects of the human motion simulator are not novel, we will focus on the schedule generation aspect in this work. In the following subsections, we discuss the technical contribution in detail.

\subsection{Problem Formulation}
This section will describe the schedule generation algorithm. Let $ST$ denote a schedule template consisting of an ordered sequence of $n$ entries, where $ST=\{st_1, st_2, ..., st_n\}$, and each entry $st_i = (a_i,t_{start}^i, v_{start}^i, d^i, v_d^i)$, containing an activity $a_i$, a start time $t_{start}^i$ with variability $v_{start}^i$, and duration $d^i$ with variability $v_d^i$. For each entry, $a_i$ is required, but $(t_{start}^i,v_{start}^i)$ may be defined or left empty, and $(d^i,v_d^i)$ may be defined or left empty.

Next, let $S$ denote a generated schedule consisting of an ordered sequence of $n$ entries, where $S=\{s_1, s_2, ... , s_n\}$, and each entry $s_i = (a_i, t_{start}^i,d^i, t_{end}^i)$, containing an activity $a_i$,  start time $t_{start}^i$, duration $d^i$, and end time $t_{end}^i$.

\subsection{Iterative Bi-directional Constraint Propagation}
Algorithm \ref{alg:schedule} describes our method for generating a concrete schedule instance $S$ based on a schedule template $ST$.

\textbf{1. Initializing Random Values} The first step is to initialize $S$ from the constraints provided by the designer in the template. In this step, random variation is added to ($t_{start}$, $d$) from uniform distributions bounded by the specified ranges ($\pm v_{start}$, $\pm v_d$). For each entry $s_i$, $t_{end}$ is initialized to $null$.

The next step is to iteratively populate any uninitialized start and end times in $S$ by alternating between two update steps: applying duration constraints, and applying adjacency constraints. Rules are also applied to eliminate gaps in the schedule and avoid time conflicts.

\begin{algorithm}
\caption{Schedule Generation Algorithm}
\label{alg:schedule}
\hspace*{\algorithmicindent} \textbf{Input} Schedule template $ST$ \\
\hspace*{\algorithmicindent} \textbf{Output} Schedule $S$
\begin{algorithmic}[1]
\Procedure{GenerateSchedule}{ST}

    \Comment{Step 1: Apply random initialization to activities}
    \For{each constraint $st_i \in ST$}
        \State $s_i[a] \gets st_i[a]$        \If{$\exists(st_i[t_{start}],st_i[v_{start}]$)}
            \State $\Delta t \gets \mathcal{U}(-st_i[v_{start}],st_i[v_{start}])$
            \State $s_i[t_{start}] \gets st_i[t_{start}] + \Delta t$
        \EndIf
        \If{$\exists(st_i[d],st_i[v_d]$)}
            \State $\Delta t \gets \mathcal{U}(-st_i[v_d],st_i[v_d])$
            \State $s_i[d] \gets st_i[d] + \Delta t$
        \EndIf
    \EndFor

    \Comment{Step 2: Iteratively propagate constraints in $S$}

    \While {$\exists s_i \ni \nexists s_i[t_{start}] \lor \nexists s_i[t_{end}]$} 
        \For{each entry $s_i \in S$}
            \State $\Call{ApplyDurationConstraints}{s_i}$
        \EndFor
        \For{each entry $s_i \in S$}
            \State $\Call{ApplyAdjacencyConstraints}{s_i}$
        \EndFor
        \If{no entries were updated in either step}
            \State \Call{Fail}{$underconstrained$}
        \EndIf
    \EndWhile
\EndProcedure
\end{algorithmic}
\end{algorithm}

\textbf{2. Solving Duration Constraints} (Algorithm \ref{alg:duration})
Because $s_i[t_{start}] + s_i[d] = s_i[t_{end}]$, we can determine the end time of any $s_i$ which has start time and duration defined, or the start time for any $s_i$ with end time and duration defined. 

After each duration update, there may be a time conflict with an adjacent or non-contiguous activity (as there may be uninitialized schedule entries lying between two time-conflicted activities). Our algorithm looks ahead to identify these conflicts at the time a duration constraint is applied.

For example, if $s_i[t_{end}] > s_j[t_{start}] \ni j > i$, then we shorten $s_i$ to end at $s_j[t_{start}]$. Any activities between them will be reduced to zero length and effectively deleted.  This procedure is applied both forwards and backwards.

Arguably, such a conflict could be treated as a schedule resolution failure, but we considered it realistic to simulate a situation where a user ``didn't get around to something" and skipped some activities to meet an upcoming time constraint.

Uninitialized start and end times are updated in this way for all applicable $s_i$.

\begin{algorithm}
\caption{Solving duration constraints}
\label{alg:duration}
\begin{algorithmic}[1]
\Procedure{ApplyDurationConstraints}{$s_i$}
    \If{$\exists s_i[t_{start}] \land \exists s_i[d] \land \nexists s_i[t_{end}]$}
        \State $j=arg\,min\{s_j[t_{start}]$ where $j > i\}$
        \State $s_i[t_{end}] \gets \min{(s_i[t_{start}] + s_i[d], s_j[t_{start}])}$
    \ElsIf{$\nexists s_i[t_{start}] \land \exists s_i[d] \land \exists s_i[t_{end}]$}
        \State $j=arg\,max\{s_j[t_{end}]$ where $j < i\}$
        \State $s_i[t_{start}] \gets \max{(s_i[t_{end}] - s_i[d],s_j[t_{end}])}$
    \EndIf
\EndProcedure
\end{algorithmic}
\end{algorithm}

\textbf{3. Solving Adjacency Constraints} (Algorithm \ref{alg:adjacency})
The next step is to update time constraints between adjacent activities. Unlike the discrete activities recorded in the datasets, which often have gaps between them, activities in our model represent persistent spatial locations, so we do not allow gaps between activities. Activities with uninitialized start times are thus defined to begin when the previous task ends, and activities with uninitialized end times conclude at the start time of the next activity. If a gap exists between two activities, the earlier activity is extended to end at the start time of the later activity.

\begin{algorithm}
\caption{Solving adjacency constraints}
\label{alg:adjacency}
\begin{algorithmic}[1]
\Procedure{ApplyAdjacencyConstraints}{$s_i$}
    \If{$\exists s_i[t_{end}] \land \exists s_{i+1}$}
        \If{$\nexists s_{i+1}[t_{start}]$}
            \State $s_{i+1}[t_{start}] \gets s_i[t_{end}]$
        \ElsIf{$s_{i+1}[t_{start}] > s_i[t_{end}]$}
            \State Extend current activity to fill gap
            \State $s_i[t_{end}] \gets s_{i+1}[t_{start}]$
        \ElsIf{$s_{i+1}[t_{start}] < s_i[t_{end}]$}
            \State \Call{Fail}{$overconstrained$}
        \EndIf
    \EndIf
    \If{$\exists s_i[t_{start}] \land \exists s_{i-1}$}
        \If{$\nexists s_{i-1}[t_{end}]$}
            \State $s_{i-1}[t_{end}] \gets s_i[t_{start}]$
        \ElsIf{$s_{i-1}[t_{end}] < s_i[t_{start}]$}
            \State Extend previous activity to fill gap
            \State $s_{i-1}[t_{end}] \gets s_i[t_{start}]$
        \ElsIf{$s_{i-1}[t_{end}] > s_i[t_{start}]$}
            \State \Call{Fail}{$overconstrained$}
        \EndIf
    \EndIf
\EndProcedure
\end{algorithmic}
\end{algorithm}

\textbf{4. Exit criteria}
The iteration of constraint resolution continues until there are no schedule entries left with unresolved start or end times, indicating success.

The algorithm can fail to meet this condition if the schedule template is underconstrained. Such cases can be detected if no updates have been made to the schedule after a round of applying both duration and adjacency constraints. In these cases, the algorithm fails and no schedule is produced.

In other cases, the schedule may be overconstrained.  That is, a time conflict between two schedule entries cannot be resolved by adjusting a single start or end time.  Rather than implementing complex logic to handle these cases, we chose to allow the algorithm to fail in these situations.

\subsection{Validation of Schedule Template}
Because failure of the system to generate a valid schedule would be problematic at run time, such as during automated testing, we provide a design-time tool that designers can use to validate the schedule templates they create. Our validation algorithm is able to identify the following errors:
\begin{itemize}
    \item Chronological errors in the template
    \item Under-constrained conditions where the time constraints are insufficient to define a complete schedule 
    \item Over-constrained conditions where overlapping activities result in unresolvable time conflicts 
\end{itemize}

Because the process of schedule generation incorporates randomness, our validation algorithm generates schedules at the minimum and maximum extremes of the variance ranges. If it can be confirmed that there are no overlapping activities or underconstrained situations given these boundary values, then the template is considered valid. These validation steps are especially important for preventing failures during automated operations like regression testing or generation of large-scale synthetic datasets of user schedules. 

\subsection{Integration with the robot simulator}
As indicated in Fig. \ref{fig:system-diagram-1}, the generated schedules were used as one input into Astro's simulator, along with a floorplan and a mapping of activities to locations.
The simulations were executed in a photo-realistic simulator integrated with Gazebo, with the human simulation module moving each user to the target destination using an A* path planner with local obstacle avoidance for their scheduled activities based on the simulation wall-clock time.

The resultant simulation can be used in two ways.  First, the simulator can be used online with the robot software, to test and evaluate robot behaviors executed in relation to the simulated user(s). Alternatively, the output can be captured as a rosbag and added to a test corpus for offline replay.

\section{Validation: Use Case Scenarios}
\begin{figure}[thb]
    \centering
    \includegraphics[width=0.95\linewidth]{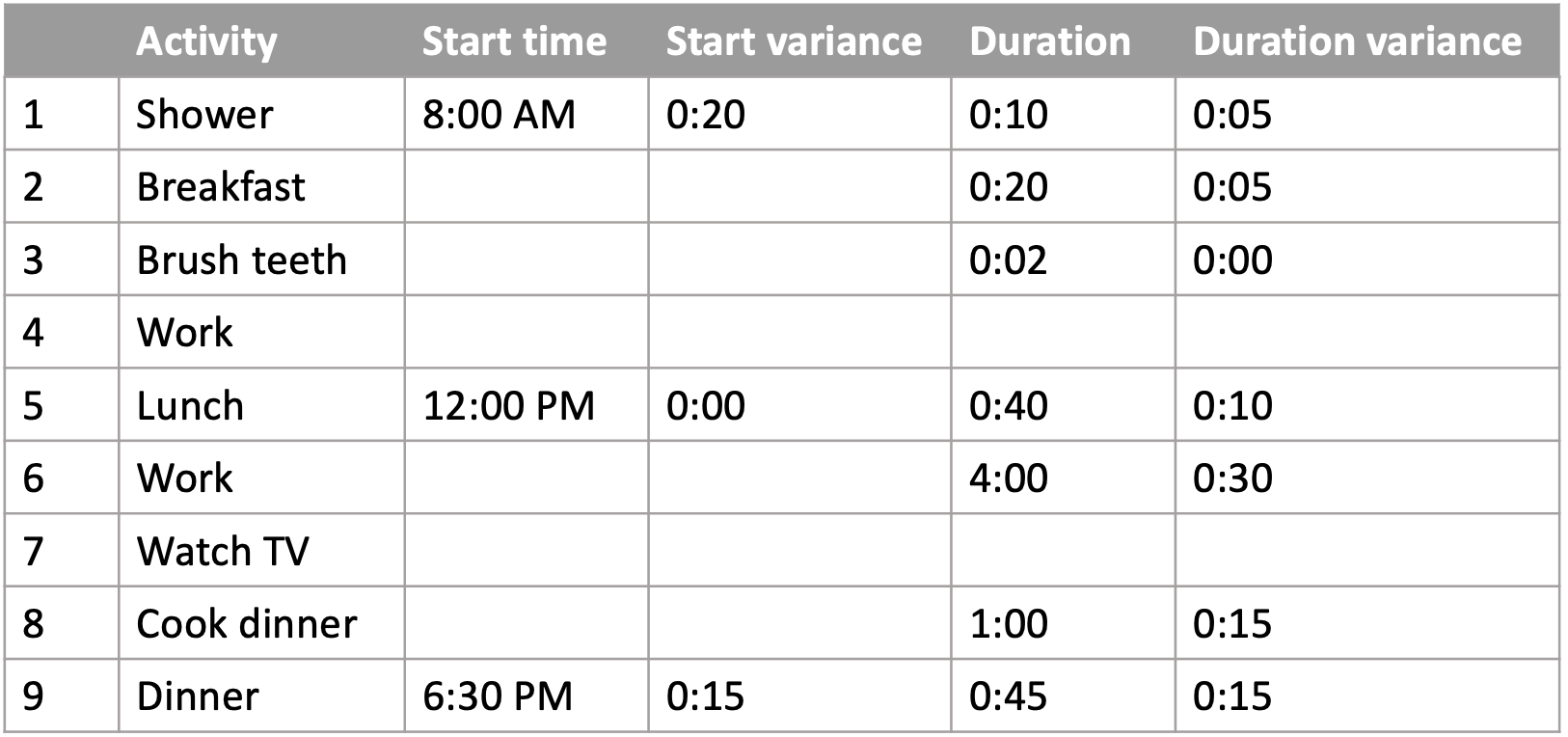}
    \caption{Example schedule template illustrating use case scenarios. Start variance, duration, and duration variance are expressed in hours and minutes.}
    \label{fig:schedule-example}
\end{figure}

To demonstrate the expressive power of the framework, we present a set of use cases representing common schedule patterns found in daily life and describe how they can be represented in a schedule template using our framework, with examples shown in Fig. \ref{fig:schedule-example}. Where applicable, we also note examples observed in the datasets studied later in Sec. \ref{sec:datasets}.


\subsection{Scheduled activities}

Some activities, like scheduled meetings and appointments, have fixed start times. In the datasets, aside from wakeup times, rigidly-scheduled activities were almost never seen. Much more common were loosely-scheduled activities where start times varying within a range. These activities, such as meal times, might be tentatively planned for a specific start time, but with the expectation that it may vary significantly.

The proposed framework supports this flexibility by allowing variance to be specified for the activity’s start time. In Fig. \ref{fig:schedule-example}, rows 1 and 9 show activities with variable start times, whereas row 5 is rigidly scheduled for 12:00 PM.

\subsection{Variable-length activities}

While some activities observed in the datasets, like brushing teeth or taking medicine, were fairly fixed in duration, many daily activities required a variable length of time to finish. This was observed in the datasets for activities such as meals and computer work. 
In the proposed framework, the duration variance can be used to control this variability. In Fig. \ref{fig:schedule-example}, ``brush teeth" (row 3) is fixed-duration, but most other entries have nonzero duration variance.

\subsection{Sequences of activities}

It is also common for activities to follow each other in a sequence.  For example, every public dataset we examined included a sequential morning routine, such as showering, breakfast, and then brushing teeth. Each of these typically followed the previous activity rather than varying independently.  In such cases, it does not make sense to express start times explicitly, but rather to chain together activities.

In the proposed framework, this can be achieved by setting the start time for the first activity, but specifying only duration and duration variance for the following activities. An example of this is shown in rows 1-3 in Fig. \ref{fig:schedule-example}.

\subsection{End time constraints}

Sometimes an activity is constrained to a fixed end time.  For example, preparation for a meal may need to be completed by a certain time, or a morning routine may need to be finished by the time a person needs to leave for work. 

This type of pattern is made possible by the backward-propagation aspect of the algorithm.  For example, consider a ``cooking" activity followed by a ``dinner" activity (\textit{e.g.} rows 8-9 of Fig. \ref{fig:schedule-example}).  If ``dinner" is given a fixed start time of 6:30pm and ``cooking" is specified to have a duration of 1 hour but no fixed start time, then the end time of ``cooking" would be constrained to the start of ``dinner" and its start time would be pushed back to 5:30pm to fit the constraint.

\subsection{Filling the time available}

In other situations, an activity can be extremely flexible in its duration, expanding to fill the time available. Activities like watching TV or reading can fall into this category.

This kind of activity can be expressed by leaving both start time and duration blank, allowing the propagated constraints from other activities to implicitly define start and end times for these flexible activities, \textit{e.g.} in rows 4 and 7 of Fig. \ref{fig:schedule-example}.

\subsection{Comparisons with other approaches}

There are many ways to specify schedules for simulations, and there are benefits and drawbacks to any method.  We believe that our method has some advantages over other approaches, as follows.  

A naive approach might specify exact start and stop times for tasks. Our approach provides control over variance of start times and durations, an important consideration for data which will be used to train machine learning systems, where variability in the distribution of data is needed for robustness.

Some approaches use programmatic logic to control task scheduling.  While such approaches are extremely flexible, they are also complex. The method proposed here contains a small set of parameters which are still sufficient to express the task categories listed above. Arguably, this can be simpler for designers to manage, tune, and debug than a representation based on complex sets of rules and conditions.

Other approaches are stochastic, e.g. using probability distributions to specify when tasks should be scheduled.  One drawback of such approaches is that it is difficult to ensure the sequential execution of dependent sequences of activities, which our proposed framework can provide.

\section{ Evaluation: Emulating captured data }
Although our proposed method is primarily intended for developing handcrafted schedules, we also evaluated how effectively it could be used to create schedule data in imitation of an example dataset.

For this evaluation, we performed comparisons with several public datasets as well as one we captured on our own, evaluating how closely we were able to emulate the activity patterns in those datasets using our framework by measuring similarity between the schedules generated by our template and schedules from the original dataset.

\subsection{Levenshtein distance as a similarity metric} To measure similarity between sequences, we used a similarity metric based on Levenshtein distance. Similar techniques have often been used to make quantitative comparisons between human motion trajectories \cite{kanda2009abstracting, calderara2011detecting, hanheide2012analysis, syaekhoni2018analyzing}.

In our comparison, we first discretize each of the sequences to be compared into a state-chain representation, where states are sampled at regular 1-minute intervals and assigned the value of the action being performed at that time step. The Levenshtein distance, denoted by $L(S_1, S_2)$, between action state sequences $S_1$ and $S_2$ of length $n$ is calculated as the minimum number of single-element edits required to transform $S_1$ into $S_2$. 

A measure of similarity between the state chains can be obtained by normalizing the Levenshtein distance and subtracting it from 1, as follows:

$$sim_{Lev}(S_1, S_2) = 1 - \frac{L(S_1, S_2)}{n}$$

\subsection{Comparison method}\label{sec:comparisons}
The datasets being compared consisted of collections of schedules, each containing several examples of daily schedules collected for one user. To compare \textbf{cross-similarity} between two collections of daily schedules, $C_1$ containing $m$ schedule examples and $C_2$ with $n$ examples, pairwise comparisons using $sim_{Lev}$ were performed between the individual schedules, and the results were averaged.

$$sim_{cross}(C_1, C_2) = \frac{\sum_{i=1}^{m} \sum_{j=1}^{n} sim_{Lev}(C_1^i, C_2^j)}{m\cdot n}$$

To help interpret the meaning of these similarity values, \textbf{self-similarity} scores were also computed within each collection of schedules. The self-similarity metric provides an approximate (although not strict) upper bound to the similarity that could be achieved when attempting to imitate a given collection of schedules. For a collection of schedules $C$ containing $n$ examples, average self-similarity over all combinations was computed as follows.

$$sim_{self}(C) = \frac{2}{n(n-1)}\sum_{1 \leq i < j \leq n}sim_{Lev}(C^i,C^j)$$

\subsection{Datasets for comparison} \label{sec:datasets}
We evaluated the ability of our framework to emulate examples from public datasets of daily activity schedules, as well as a dataset of space transitions in daily activity that we captured ourselves.

\subsubsection{Public datasets} 
The following three public datasets were used in this comparison. When necessary, an activity designated ``other" was added to fill gaps between activities.

\begin{figure*}[t]
    \centering
    \includegraphics[width=\textwidth]{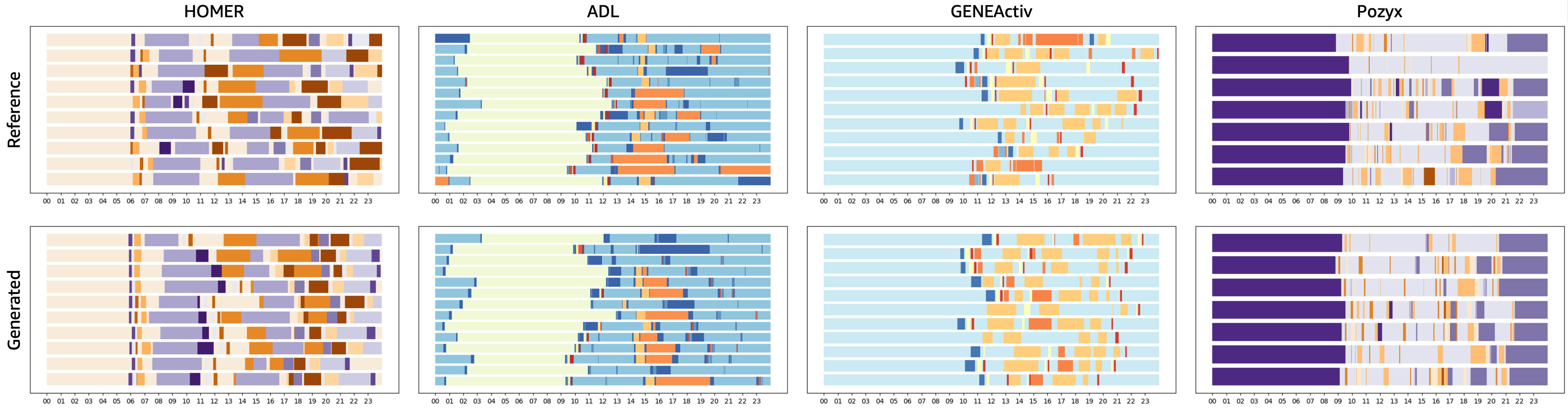}
    \caption{Results of manual fitting of schedule templates to reference datasets. Each graph shows activities over a 24-hour period (activity names omitted for readability). The upper graph of each pair represents the original reference data, and the lower graph shows data generated by our algorithm.}\label{fig:fitting-results}
\end{figure*}

\textbf{HOMER Dataset:} This dataset includes simulated activity schedules generated based on self-reported schedule information collected via Amazon Mechanical Turk. We used the ``test" data for household 0, consisting of 12 activities tracked over 10 days. \cite{patel2022proactive}

\textbf{ADL Dataset:} A dataset containing manually-labeled activities of daily living performed by two users on a daily basis in their own homes. We used data from user 1, comprising 10 activities tracked over 14 days. \cite{ordonez2013activity}

\textbf{GENEActiv Dataset:} This is a smartwatch accelerometer dataset. We used the manually-annotated activities for user 1, featuring 7 activities tracked over 11 days. \cite{garcia2014long}

\subsubsection{Self-collected data}
The above public datasets tracked daily activities, but reported activities were sparse and did not exactly correspond to locations. For our robot applications, our primary interest is in activities representing location changes within the home. Thus, we collected an additional dataset ourselves, tracking a single participant (one of the authors of this paper) across 7 discrete regions of the home (8 states, including ``not tracked"), over 7 days of daily activity.

\textbf{Pozyx Dataset}: In this dataset, the participant’s real-time positioning data were collected using the Pozyx Real-Time Localizing System\footnote{\url{https://www.pozyx.io/}}, which performs position tracking of portable ultra-wideband radio emitter tags worn by participants using the Time Differences of Arrival (TDoA) method with multiple fixed antennas. For each day of data collection, the participant started the tracking system between 9-10am and the positioning data were continuously generated for a collection window of 11 hours at frequency of 1Hz. The participant's location was tracked at room-level resolution, and we applied a low-pass filter to remove noisy space transitions with durations less than 1 minute.

As it is synthetically generated, the HOMER data is the least realistic dataset in our study. The GENEActiv and ADL datasets represent real activity data, but it was manually reported and only captures a few discrete actions. The Pozyx data represents raw space transitions in the home, so it is the noisiest data; however, it most closely represents the kind of data we plan to use with our framework.

\subsection{Experiment}
For each dataset, we manually created a schedule template to emulate the behavior patterns with our framework.

For each reference dataset to be evaluated, we compared three conditions: a random \textbf{baseline} (10 schedules), the cross-similarity score between the \textbf{generated} data (10 schedules) and reference data (7-14 schedules), and the \textbf{self-similarity} score within the reference data. 

\subsubsection{Baseline} As a baseline for this comparison, we created an algorithm that triggers a random user activity every 30 minutes. Although this is a naive approach, it is not unreasonable as a first-pass design for a simulator to run in a newly-designed environment when models of user behavior are not available.

\subsubsection{Generated schedules} For this comparison, we manually created schedule templates designed to imitate the target datasets as closely as possible. Although an automated method for doing this would be useful, we leave that for future work. The goal of this evaluation is to demonstrate the expressive power of the proposed framework, which can be done via manual fitting.

For activities occurring around consistent times, $t_{start}$, $v_{start}$, $d$, and $v_d$ were defined. For sequences of transient activities which seemed to be more dependent on each other than on a fixed time, only $d$ and $v_d$ were defined. Often, ``default" states such as \textit{other} or \textit{Spare time} filled the time between other activities, so these were specified with all constraint fields empty, allowing them to fill the time between other activities. Intermittent activities were modeled by defining $v_d > d$, as this would cause the duration to sometimes go below zero, effectively deleting the activity on some days. Finally, templates were iteratively refined until the generated data appeared similar to the reference data.

\subsubsection{Self-similarity} Whereas the baseline provided a practical lower bound for similarity measurements, we approximated an upper bound by computing a self-similarity score for each dataset as described in Sec. \ref{sec:comparisons}, representing the degree of day-to-day variation in that dataset.

\subsection{Results}

Fig. \ref{fig:fitting-results} shows the results of our manual fit to each of the four datasets. This visualization was a helpful tool during the development of the templates, and it provides a broad sense of the degree of similarity which was achieved. 

\begin{figure}[thbp]
    \centering
    \includegraphics[width=\linewidth]{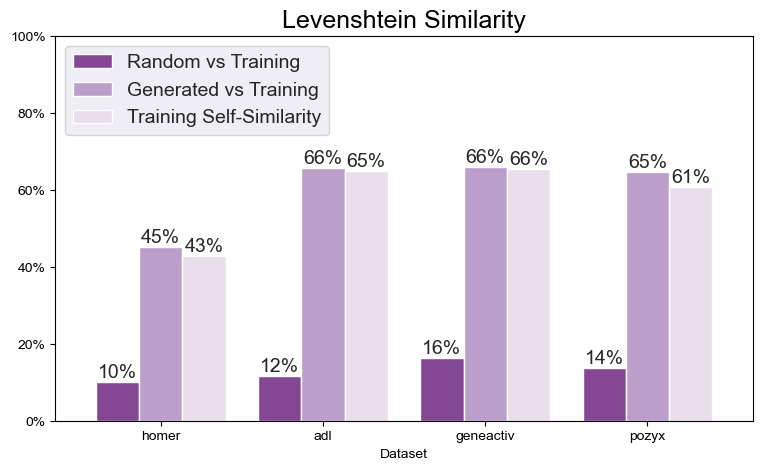}
    \caption{Similarity scores for each dataset using the Levenshtein distance-based similarity metric, where higher similarity indicates a closer approximation of the real schedule. In all cases the similarity between the generated data and reference data was much higher than the random baseline, and comparable to the self-similarity within the reference set itself.}
    \label{fig:levenshtein}
\end{figure}

The quantitative similarity scores are shown in Fig. \ref{fig:levenshtein}. In all cases, the similarity between the generated data and reference data was much higher than the random baseline, and comparable to the self-similarity within the reference set itself. The best match was to the GENEActiv dataset, with a 66.0\% similarity score, slightly higher than the 65.6\% reference data self-similarity and much better than the 16.4\% similarity in the random baseline case.  The match with lowest similarity was with the HOMER dataset, although the 45.3\% similarity still beat the 43.0\% self-similarity within that dataset and was much higher than the 10.2\% similarity with the baseline case. The greatest improvement over the baseline was observed in the ADL dataset, with 65.9\% similarity vs 11.9\% in the baseline. From these results we conclude that the expressive power of the proposed framework was sufficient to closely approximate each of these datasets.

Cases where the cross-similarity scores exceeded the self-similarity scores likely indicate that the generated schedules had lower variability than the reference schedules and were hence closer to the ``average" schedule than the original data. This is expected, as the original schedules included day-to-day variations of activity sequence that could not be reproduced using our method.

Qualitatively, we visually judged the generated sequences to closely resemble the reference data. Some anomalies in the reference data were not reproducible, but overall patterns appeared highly similar to the reference data in three key ways: First, several activity sequences such as morning and bedtime routines were successfully reproduced. Second, the lengths of individual activities and the relative distributions of activities through the day were roughly accurate. Finally, activity selection appeared to happen at the correct times of day and in the right sequence, with things like showers, meal times, commutes, and taking medication (activities tracked in the public datasets) occurring around the correct times.

\section{Discussion}
\subsection{Fitting Templates to Data}
For the most part, it was not difficult to develop schedule templates to approximate a given input dataset.  For activities which seemed to be associated with specific times of the day, such as waking up or eating meals, we used absolute start times with some variance. For activity sequences, we omitted start times and specified only duration and duration variance. For activities which occurred on some days but not others, we set the duration variance to be greater than the duration, resulting in the activity occasionally having zero duration (although this is hacky - it is not possible to model activities of fixed duration in this way).
However, despite this flexibility, some phenomena were still difficult to model, such as the following examples.

\subsubsection{Activity pattern changes} In the ADL dataset, the user ate lunch at home on some days, but went out for lunch on others. As noted above, our framework can handle a single activity occasionally being omitted, but it has no way to model a switch like that, where the ``lunch" activity is sometimes replaced by a ``leaving" activity with an earlier start time, and the two never occur together.  In such cases, one possible solution could be to create multiple templates for different daily patterns, and have the schedule generator select randomly among them.

\subsubsection{Stochastic vs scheduled activities} The Pozyx dataset was the most difficult for us to fit.  In part this is because much of the activity appeared to be driven by random events, like going to the restroom or getting a glass of water, rather than scheduled activities that occur at a certain time or in a certain sequence.  Rather than trying to account for such random events within a calendar schedule, we believe it could make sense to model them as a parallel phenomenon. The relative balance between scheduled and stochastic events seems likely to depend the individual user's lifestyle.

\subsection{Limitations and Future Work}
\subsubsection{Similarity metrics} One limitation of using Levenshtein distance is the fact that it is based on total number of insertions or deletions. This means that this metric will have less sensitivity to short-duration activities than to longer-running activities, which can bias the scoring.  However, we observed that as we improved our hand-tuned schedule templates over time, the similarity metric indeed converged towards the self-similarity score, indicating that the directionality of the metric was consistent with what we were evaluating, even if there might be some bias in its absolute magnitude. 

Although the Levenshtein distance metric was included to provide a quantitative evaluation for this paper, in practice we found the graphical plots of daily schedules to provide much more useful feedback about the quality and nature of the fit, both for tuning the schedule templates and for validating the quality of the match through qualitative inspection.

\subsubsection{Automating template fit} We expect that this template fitting process could be automated, but it is not trivial to do so, due to the insertion and removal of discrete schedule entries, the existence of optional fields, and the asymmetric nature of activities where the duration variance exceeds the duration. Due to these complications, we leave automation of this process for future work.

\subsubsection{Simulating interaction between robot and human} As the data generated by this framework are meant for robot testing, one of the most important considerations is how to simulate interactions between users and the robot.  Our view is that human-robot interactions need to be modeled separately in the simulator. The proposed framework can produce realistic locations and activity contexts for a simulated human, but any simulation of interaction with the robot needs to explicitly account for the user's motivation for interacting and what behaviors they will perform - for example, will the user only interact when the robot is in the same room, or will they leave the room to search for the robot? 

\subsubsection{Multiple users} The framework was designed to implicitly support multiple simulated users.  By specifying unique locations for each user's activities, such as assigning each user a different seat in the dining room or living room, the users can be simulated to coexist in the same home. However, the random variation added to their activity schedules is applied independently, meaning that activities cannot really be coordinated between the simulated users.  For simulating situations such as a family coming together for a meal, this may be important to address in the future.

\subsection{Differences between academia and industry}

\subsubsection{Intellectual property} We were unfortunately unable to share the code of our simulation framework as open-source, due to its proprietary nature. While perhaps more common in industry, such situations can also occur in academia, depending on funding sponsors or licensing restrictions. Despite these constraints, we believe it is still possible to share useful knowledge and findings.
\subsubsection{Primary use cases} While academia rewards ``interesting" research topics, research work in industry is driven by business goals. The framework presented here was developed for manual design and adjustment of schedule templates for testing. If this were a purely academic paper, it might be more likely to focus on automated or learning-based methods of generating schedule templates, which are interesting topics but not necessarily driven by such practical needs.  Thus, we can probably expect research work from industry to trend towards the pragmatic, moreso than the theoretically interesting. This difference could be beneficial, introducing a useful diversity of perspectives into the academic discussion.

\subsubsection{Numerical performance vs practical utility} In a typical HRI study, there would be a focus on p-values and human subjects experiments. From an industry perspective, our main focus is on understanding the practical utility and expressive capability of the framework to model a variety of scenarios, and we seek to understand its strengths and limitations rather than focusing on numerical results. 

\section{Conclusion}
In this work we have introduced a framework for simulating daily user activity at scale for testing and developing commercial robots. The framework enables manual tuning of daily schedules for developing both typical-use and corner-case tests for testing and quality assurance. Our case studies demonstrated the framework's expressive capability, and we also evaluated the approach's potential to emulate example data using both public and internally-captured datasets. 

Overall, this work makes a significant contribution to the field of social robotics by providing a systematic approach for testing robot behaviors related to daily user activity. By helping to avoid training data bias, our approach has the potential to make robot behavior effective for a broad range of user households, making it a valuable tool for future research and development in social robotics.

\section*{ACKNOWLEDGMENT}
We thank Tarun Morton, Aarthi Raveendran, and Abraham Dauhajre for their support of this research. This work was supported in part by the Office of Naval Research under Grant award N00014-21-1-2584.

\addtolength{\textheight}{-12cm}   




\bibliographystyle{plainnat}
{\footnotesize\bibliography{references}}

\end{document}